\documentclass[10pt,twocolumn,letterpaper]{article}

\usepackage{cvpr}
\usepackage{times}
\usepackage{epsfig}
\usepackage{graphicx}
\usepackage{amsmath}
\usepackage{amssymb}
\usepackage{multirow}
\usepackage{verbatim}

\usepackage[pagebackref=true,breaklinks=true,letterpaper=true,colorlinks,bookmarks=false]{hyperref}

\cvprfinalcopy 

\def\cvprPaperID{2696} 

\ifcvprfinal\fi
\begin{document}
\newcommand{\todo}[1]{{\bf To do: }{{\em #1}}\\}
\newcommand{\sign}[1]{{\mbox{sign}}{{#1}}}
\newcommand{\matx}[1]{{\bf {#1}}}
\newcommand{\plan}[1] {{\bf Plan:}\\ {\em {#1}}}
\newcommand{\norm}[1]{{\mid\!\mid\!\!{#1}\!\!\mid\!\mid}}
\newcommand{\pr}[1]{{\rm Pr}[#1]}
\renewcommand{\contentsname}{\vspace*{-1.2cm}}
\newcommand{\newterm}[1]{{\bf #1}}
\newcommand{\capsty}[1]{{\em {#1}}}
\newcommand{\vect}[1]{{\bf {#1}}}
\newcommand{\argmax}[1]{{\begin{array}{c}\mbox{argmax}\\{{#1}}\end{array}}}
\newcommand{\mean}[1]{{\begin{array}{c}\mbox{mean}\\{{#1}}\end{array}}}
\newcommand{\hussein}[1]{\textcolor{red}{#1}}
\title{Adversarial Examples that Fool Detectors}
\author{Jiajun Lu\thanks{Both authors contributed equally}, Hussein Sibai\footnotemark[1], Evan Fabry\\
University of Illinois at Urbana Champaign\\
\{jlu23, sibai2, efabry2\}@illinois.edu
}
\maketitle

\begin{abstract}
An {\em adversarial example} is an example that has been adjusted to produce a wrong label when presented to a system at test time.
To date, adversarial example constructions have been demonstrated for classifiers, but not for detectors.
If adversarial examples that could fool a detector exist, they could be used to (for example) maliciously create security hazards on roads
populated with smart vehicles.  In this paper, we demonstrate a construction that successfully fools two standard
detectors, Faster RCNN and YOLO. The existence of such examples is surprising, as attacking a classifier is very
different from attacking a detector, and that the structure of detectors -- which must search for their own  bounding
box, and which cannot estimate that box very accurately -- makes it quite likely that adversarial patterns are strongly
disrupted. We show that our construction produces adversarial examples that generalize well across sequences digitally, even though
large perturbations are needed. We also show that our construction yields physical objects that are adversarial. 
\end{abstract}

\section{Introduction}

An {\em adversarial example} is an example that has been adjusted to produce a wrong label when presented to a system
at test time. In the literature, adversarial examples with imperceivable
perturbations and unexpected properties (e.g. transferability) are one of the biggest mysteries of neural networks. 
There is a range of constructions~\cite{DBLP:journals/corr/Moosavi-Dezfooli15, nguyenfooled,
  moosavi2016universal} that yield adversarial examples for image classifiers, and 
there is good evidence that small imperceivable adjustments suffice. Furthermore, Athalye {\em et al.}~\cite{athalye2017synthesizing}
show that it is possible to build a physical object with visible perturbation patterns that is persistently misclassified by standard image classifiers
from different view angles at a roughly fixed distance.
There is good evidence that adversarial examples built for one classifier will fool others, too~\cite{papernottransfer, Song2016}.  
The success of these attacks can be
seen as a warning not to use highly non-linear feature constructions without  having strong mathematical
constraints on what these constructions do; but taking that position means one cannot use methods that are largely
accurate and effective. 

Detectors are not classifiers.  A classifier accepts an image and produces a label. In contrast, a detector, like
Faster RCNN~\cite{ren2015faster, chen2017implementation}, identifies bounding boxes that are ``worth labelling'', and
then generates labels (which might include {\tt background}) for each box. The final label generation step employs a
classifier.  However, the statistics of how bounding boxes cover objects in a detector are complex and not well understood.  Some
modern detectors like YOLO 9000~\cite{redmon2016yolo9000} predict boxes and labels using features on a fixed grid,
resulting in fairly complex sampling patterns in the space of boxes, and this means that pixels outside a box may
participate in labelling that box. Another important difference is that detectors usually have RoI pooling or feature map
resizing, which might be effective at disrupting adversarial patterns. 
To date, no successful adversarial attack on a detector has been demonstrated.
In this paper, we demonstrate successful adversarial attacks on Faster RCNN, which generalize to YOLO 9000.

We also discuss the generalization ability of adversarial examples. We say that an adversarial perturbation generalizes if, when
circumstances (digital or physical) change, the corresponding images remain adversarial. For example, a perturbation of a stop 
sign generalizes over different distances if it remains adversarial when the camera approaches the stop sign. 
An example generalizes better if it remains adversarial for more cases (e.g. changes of detector, background and lighting). 
If an adversarial example cannot generalize, it is not a threat in the majority of real world systems.

Our contributions in this paper are as follows:

\begin{itemize}
\item We demonstrate a method to construct adversarial examples that fool Faster RCNN digitally; examples produced
by our method are reliably either missed or mislabeled by the detector. These examples without modification also fool 
YOLO 9000, indicating that the construction produces examples that can transfer across models.
\item Our adversarial examples can be physically created successfully, and they still can fool detectors in suitable circumstances. They
can also slip through  recent strong image processing defenses against adversarial examples.
\item In practice, we find that adversarial examples require quite large disruptions to the pattern on the object in order to fool detectors.
Physical adversarial examples require bigger disruptions than digital examples to succeed. 

\end{itemize}

\section{Background}

Adversarial examples are of interest mainly because the adjustments required seem to be very small and are
easy to obtain~\cite{szegedy2013intriguing, goodfellow2014explaining, DBLP:journals/corr/FawziMF16}.  Numerous search
procedures generate adversarial examples~\cite{DBLP:journals/corr/Moosavi-Dezfooli15, nguyenfooled,
  moosavi2016universal}; all searches look for an example that is (a) ``near'' a correctly labelled example
(typically in $L_1$ or $L_2$ norm), and (b) mislabelled.  Printing adversarial images then photographing them can retain their adversarial
property~\cite{DBLP:journals/corr/KurakinGB16, athalye2017synthesizing}, which suggests that adversarial examples might exist
in the physical world.  Their existence could cause a great deal of mischief.   
There is some evidence that it is difficult to build physical examples that fool a stop sign detector~\cite{lu2017no}.  In
particular, if one actually takes a video of an existing adversarial stop sign, the adversarial pattern does not
appear to affect the performance of the detector by much.  Lu {\em et al.} speculated that this might be because adversarial
patterns were disrupted by being viewed at different scales, rotations, and orientations.  This created some
discussion.  OpenAI demonstrated a search procedure that could produce an image of a cat that was misclassified  when
viewed at multiple scales~\cite{athalye2017synthesizing}.  There is some blurring of the fur texture on the cat they generate, but
this would likely be imperceptible to most observers.  OpenAI also demonstrated
an adversarial image of a cat that was misclassified when viewed at multiple scales {\em and}
orientations~\cite{athalye2017synthesizing}.  However, there are significant visible artifacts on that image; few would
think that it had not been obviously tampered with.

Recent work has demonstrated physical objects that are persistently misclassified from
different angles at a roughly fixed distance~\cite{athalye2017synthesizing}. The search procedure
manipulates the texture map ${\cal T}$ of the object.  The procedure samples a set of viewing conditions ${\cal V}_i$
for the object, then renders to obtain images $I({\cal V}_i, {\cal T})$.  Finally, the procedure adjusts the texture map
to obtain images that are (a) close to the original images and (b) have high probability of misclassification.  The
adversarial properties of the resulting objects are robust to the inevitable errors in color, etc., in
producing physical objects from digital representations.

{\bf Defences:}  There is fair evidence that it is hard to tell whether an example is adversarial (and so (a)
evidence of an attack and (b) likely to be misclassified) or not~\cite{shaham2015understanding,DBLP:journals/corr/GuR14,
  Sharif:2016:ACR:2976749.2978392,metzen2017detecting,carlini2016defensive,fawzi2015analysis}.    
Current procedures to build adversarial examples for deep networks appear
to subvert the feature construction implemented by the network to produce odd patterns of activation in late stage
ReLU's; this can be exploited to build one form of defence~\cite{lu2017safetynet}.   There is some evidence that other
feature constructions admit adversarial attacks, too~\cite{metzen2017detecting}.  However, adversarial attacks typically
introduce unnatural (if small) patterns into images, and image processing methods that remove such patterns yield
successful defenses. Guo {\em et al.} showed that cropping and rescaling, bit depth reduction, JPEG compression and
decompression, resampling and reconstructing using total variation criteria, and image quilting all provide quite
effective ways of removing adversarial patterns~\cite{guo2017countering}.

{\bf Detectors and classifiers:} It is usual to attack classifiers, and all the attacks we are aware of
attack on classifiers.  However, for many applications, classifiers are not useful by themselves.  Road sign is a good
example.  A road sign classifier would be applied to images that consist largely of a road sign (e.g. those
of~\cite{stallkamp2012man}).  But there is little application need for a road sign classifier except as a 
component of a road sign detector, because it is unusual in practice to deal with images that consist largely of road
sign. Instead, one usually deals with images that contain many things, and must find and label the road sign. It is
natural to study road sign classifiers (e.g.~\cite{sermanet2011traffic}) because image classification remains difficult and academic
studies of feature constructions are important.  But there is no particular threat posed by an attack on a road sign
classifier.  An attack on a road sign detector is an entirely different matter.  For example, imagine the danger if one could get a
template that, with a can of spray paint, could ensure that a detector reads a stop sign as a yield sign (or worse!). As
a result, it is important to know whether (a) such examples could exist and (b) how robust their adversarial property is in practice.

\begin{figure}
	\centerline{\includegraphics[width=1.0\linewidth]{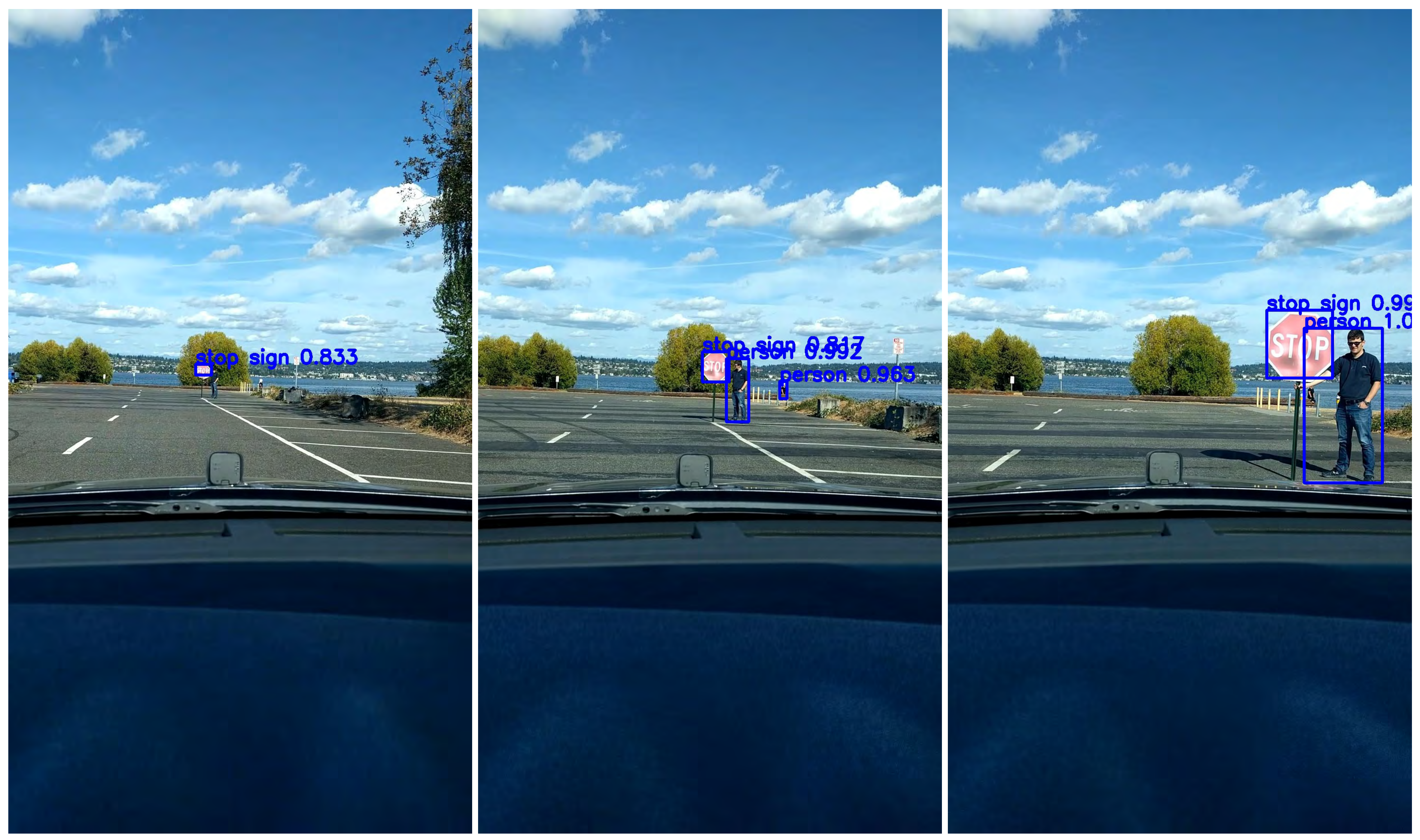}}
	\caption{Evtimov {\em et al.}~\cite{evtimov2017robust} generated physical stop signs that are misclassified, however, 
these stop signs are reliably detected by Faster RCNN. Images from Figure 10 in ~\cite{lu2017standard}. }
	\label{fig:cmp}
\end{figure}

Recently, Evtimov {\em et al.} have shown several physical stop signs that are
misclassified~\cite{evtimov2017robust}. They cropped the stop signs from the frames before presenting them to the classifier.
By cropping, they have proxied the box-prediction process in a detector; however, their
attack is not intended as an attack on a detector (the paper does not use the word ``detector'', for example).   Lu et
al. showed that their construction does not fool a standard detector~\cite{lu2017standard}, likely because the cropping process
does not proxy a detector's box selection well, and suggested that constructing an
adversarial example that fools a detector might be hard. Figure~\ref{fig:cmp} shows their stop signs presented in~\cite{evtimov2017robust} 
are reliably detected by Faster RCNN. 

\section{Method}

We propose a method to generate digital and physical adversarial examples that are robust to changes of viewing conditions.
Our registration and reconstruction based approach generates adversarial perturbations from video sequences of an object with 
moving cameras. We require the objects in the videos to be accurately aligned in 3D space. We can easily register stop signs as they are 2D polygons. 
Moreover, we can accurately register face images to a virtual 3D face model. Hence, we perform our experiments on these two types of data. 

\subsection{Approach for stop signs}
\label{sec:attacking_stop_signs}
We use the stop sign example to demonstrate our attack, which extends to other objects that are 
registered from image domain to some root coordinate system (e.g. faces in section~\ref{sec:face}).
We search for an adversarial pattern that (a) looks like a stop sign and (b) fools Faster RCNN.
We select a set of $N$ diverse frames ${\cal I}_i$ as the training examples to generate the pattern.
A stop sign is represented as a texture map ${\cal T}$ in some root coordinate system. We construct
(currently by hand) correspondences between eight vertices on the stop sign instances in training frames and the vertices of ${\cal T}$.  
We use these correspondences to estimate a viewing map ${\cal M}_i$, which maps the texture ${\cal T}$ in the root coordinate system
to the appropriate pattern in training frame ${\cal I}_i$. We also incorporate in ${\cal M}_i$ the illumination intensity, which is estimated
 by computing the average intensity over the stop sign in the image. Relative illumination intensities are used to
scale the adversarial perturbations.  We write ${\cal I}({\cal M}_i, {\cal T})$ for
the image frame obtained by superimposing ${\cal T}$ on the frame ${\cal I}_i$ using the mapping ${\cal M}_i$; ${\cal
  B}_s({\cal I})$ for the set of stop sign bounding boxes obtained by applying 
Faster RCNN to the image ${\cal I}$; and $\phi_s(b)$ for the score produced by Faster RCNN's classifier for a stop sign in box
$b$. To produce an adversarial example, we minimize the mean score for a stop sign produced by Faster RCNN in all training images as a function
of ${\cal T}$, that is \[
\Phi({\cal T})=\frac{1}{N} \sum_{i=1}^N \mean{b \in {\cal B}_s({\cal I}({\cal M}_i,
  {\cal T}))} \phi_s(b)\] possibly subject to some constraint on ${\cal T}$, such as being close to a normal stop sign in $L_2$ distance. 
We have also investigated minimizing the maximum score for all the stop sign proposals, and found that minimizing the mean score 
gives slightly better results.

{\bf Minimization procedure:}  First, we compute $\nabla_{\cal T} \Phi({\cal T})$ by
computing $\nabla_{\cal I} \mean{b \in {\cal B}_s({\cal I}({\cal M}_i, {\cal T})} \phi_s(b)$. The gradients in the frame coordinate system
are mapped to the root coordinate system with inverse view mapping ${\cal M}^{-1}_i$, and then are cropped to the extent of ${\cal T}$ in 
that coordinate system. We average gradients mapped from all $N$ training frames. However, directly using the gradients to take large steps
frequently stalls the optimization process. Instead, we find that computing the descent direction with the sign of the gradients for given pattern
${\cal T}^{(n)}$ ($n$ stands for iteration number) facilitates the optimization process. 
\[
\vect{d}^{(n)}=\sign (\nabla_{\cal T} \Phi({\cal T}^{(n)})).
\]
We choose a very small step length $\epsilon$ such that $\epsilon \vect{d}^{(n)}$ represents an update of one least
significant bit, which leads to an optimization step of form
\[
{\cal T}^{(n+1)}={\cal T}^{(n)}+\epsilon \vect{d}^{(n)}.\]
The optimization process usually takes hundreds or even thousands of steps. One termination criterion is to stop the optimization
when the pattern fools the detector more than 90\% of the cases on the validation set. Another termination criterion is a fixed number of iterations. 

{\bf Why large steps are hard:} In our experiments, taking large steps with unsigned gradients stalls the optimization process, and
we believe large steps are hard to take for two reasons. First, each instance of the
pattern occurs at a different scale, meaning that there must be some up- or down-sampling of the gradients when mapped
to the root coordinate system.  Although we register the images with subpixel accuracy, and use a bilinear method to interpolate the transformation
process, signal losses are still inevitable.  In section~\ref{sec:multiple}, we show some evidence that this effect may make our patterns more, rather
than less, robust. Second, the structure of the network means that the gradient is a poor guide to the behavior of $\phi_s(b)$ over large scales.  In
particular, a ReLU network divides its input space into a very large number of cells, and values at any layer before the
softmax layer are a continuous piecewise linear function within a cell.  Because the network is trained to
have a (roughly) constant output for large pieces of its input space, the gradient must wiggle from cell to cell, and so
may be a poor guide to the long scale behavior of the function.

{\bf Constraining distance to the original stop sign:} 
In order to create less perceivable adversarial perturbations, we constrain the distance to the original
stop sign to be small. An $L_2$ distance loss is added to the cost function, and we minimize
\[
\Phi({\cal T})+\lambda \norm{{\cal T}-{\cal T}^{(0)}}^2
\]
Our experiments show that this distance constraint still cannot help to create small perturbations, but greatly changes the pattern
of the perturbations, refer to Figure~\ref{fig:adv_stop}.

We create our physical adversarial stop signs by printing the pattern ${\cal T}$, cutting out the printed stop sign area, and
sticking it to an actual stop sign (30 in by 30 in).

\subsection{Extending to faces}
\label{sec:face}
We extend the experiments onto faces, which have complex geometries and larger intra class variances, to demonstrate that our analysis
generalizes to other classes. In the face setting, we search for a pattern that fools a Faster RCNN based face detector~\cite{jiang2017face} and looks
like the original face. Our root coordinate system for faces is a virtual high quality face mesh generated from morphable face model~\cite{Blanz1999Morphable}. 
For video sequences of a face, we reconstruct the geometry of the face in the input frames using morphable face model built from the FaceWarehouse~\cite{Cao14:FaceWarehouse} data.
The model produces a 3D face mesh $F(w_i, w_e)$ that is a function of identity parameters $w_i$, and expression parameters $w_e$. 
FaceTracker~\cite{FaceTracker} is used to detect landmarks $l_i$ on the face frames, then we recover parameters and poses of the face mesh by minimizing the distances 
between the projected landmark vertices and their corresponding landmark locations on the image planes. This construction
gives us pixel-to-mesh and mesh-to-pixel dense correspondences between all face frames and the root face coordinate system (shared face mesh). 
By projecting image pixels to the face meshes via barycentric coordinates, we can achieve subpixel accurate pixel-to-pixel registrations between all face frames
(via root coordinate system). This correspondences are used to transfer the gradients from face image coordinates to the root coordinate system, then we merge the 
gradients from multiple images and reverse transfer the merged gradients back to the face image coordinates. 

\begin{figure}
	\centerline{\includegraphics[width=1.0\linewidth]{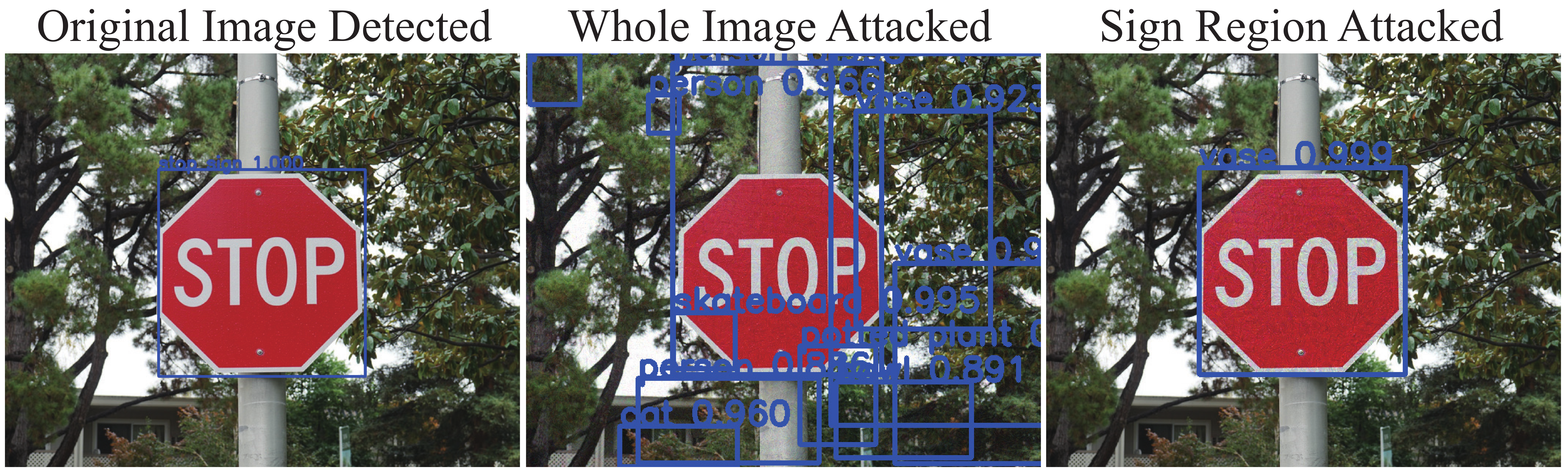}}
	\caption{Modifying a single stop sign image to attack Faster RCNN is successful. In the original stop sign image (first), the stop sign is reliably detected. In the second image,
small perturbations are added to the whole image, and the stop sign is not detected. In the last image, small perturbations are added to the stop sign region instead of 
the whole image, and the stop sign is detected as a vase.}
	\label{fig:stop_single}
\end{figure}

\begin{figure}
	\centerline{\includegraphics[width=1.0\linewidth]{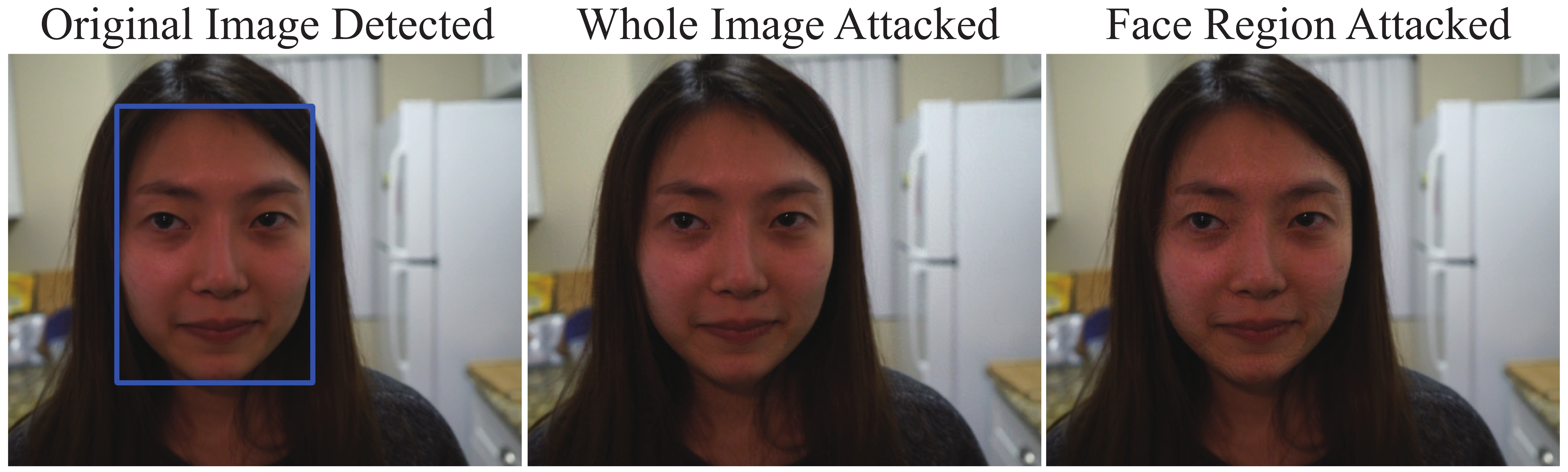}}
	\caption{Modifying a face image to attack Faster RCNN is successful. In the original face image (first), the face is reliably detected. In the second image, small perturbations
are added to the whole image, and the face is not detected. In the last image, slightly larger perturbations are added to the face region instead of the whole image, and
the face is not detected. }
	\label{fig:face_single}
\end{figure}

\begin{figure*}
	\centerline{\includegraphics[width=1.0\linewidth]{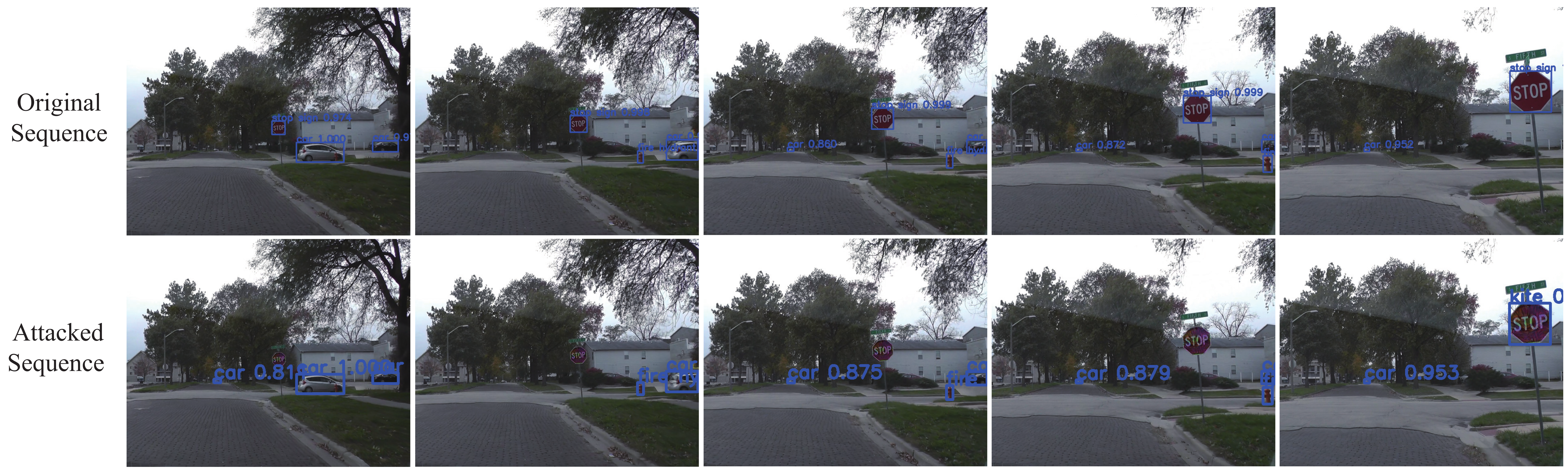}}
	\caption{Adversarial examples of stop signs for Faster RCNN that can generalize  across view conditions. The original sequence in the first row 
is a test video sequence captured for a real stop sign, and the stop sign is detected in all the frames. 
We apply our attack to a training set of videos to generate a cross view condition 
adversarial perturbation, and apply that perturbation on this test sequence to generate the attacked sequence in the second row. This is a digital attack, 
and the stop sign is either not detected or detected as a kite. }
	\label{fig:stop_multiple}
\vspace{-1ex}
\end{figure*}

\begin{figure*}
	\centerline{\includegraphics[width=1.0\linewidth]{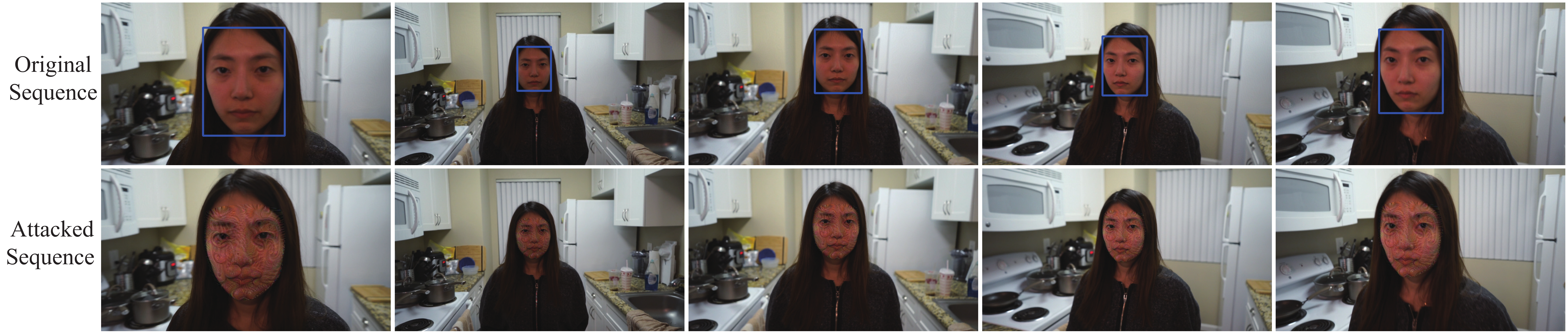}}
	\caption{Adversarial examples of faces for Faster RCNN based face detector~\cite{jiang2017face} that can generalize  across view conditions. 
The original sequence images in the first row are sampled from a test video sequence, and all the faces are reliably detected. 
We apply our attacking method to a training set of videos to generate a cross view condition 
adversarial perturbation, and apply that perturbation on this test sequence to generate the attacked sequence in the second row. This is a digital attack.}
	\label{fig:face_multiple}
\vspace{-2ex}
\end{figure*}

\section{Results}
In this section, we describe in depth the experiments we did and the results we got. Our supplementary materials include videos and more results, and it can be downloaded
from \url{http://jiajunlu.com/docs/advDetector_supp.zip}. Our high resolution paper can be downloaded from \url{http://jiajunlu.com/docs/advDetector.pdf}.
But let us first give a quick overview of our findings:
\begin{itemize}
\item Adding small perturbations suffices to fool a given object detector on a single image.
\item Enforcing the adversarial perturbations to generalize across view conditions requires significant changes to the pattern.
\item Our successful attacks for Faster RCNN generalize to YOLO.
\item Our successful attacks with very large perturbations generalize to the physical world objects in suitable circumstances.
\item Simple defenses fail to defeat adversarial examples that can generalize. 
\end{itemize}

Detectors are affected by internal thresholds. Faster RCNN uses a non maximum suppression threshold and a confidence threshold. 
For stop signs, we used the default configurations. For faces, we found the detector is too willing to detect faces, and we made it less
responsive to faces (nms from 0.15 to 0.3, and conf from 0.6 to 0.8). We used default YOLO configurations. 

\subsection{Attacking single image}
We can easily adjust a pattern on a single image to fool a detector (stop signs in Figure~\ref{fig:stop_single} and faces in 
Figure~\ref{fig:face_single}), and the change on the pattern is tiny. While this is of no practical significance, it shows our search method can
find very small adversarial perturbations. 

\begin{figure}
	\centerline{\includegraphics[width=1.0\linewidth]{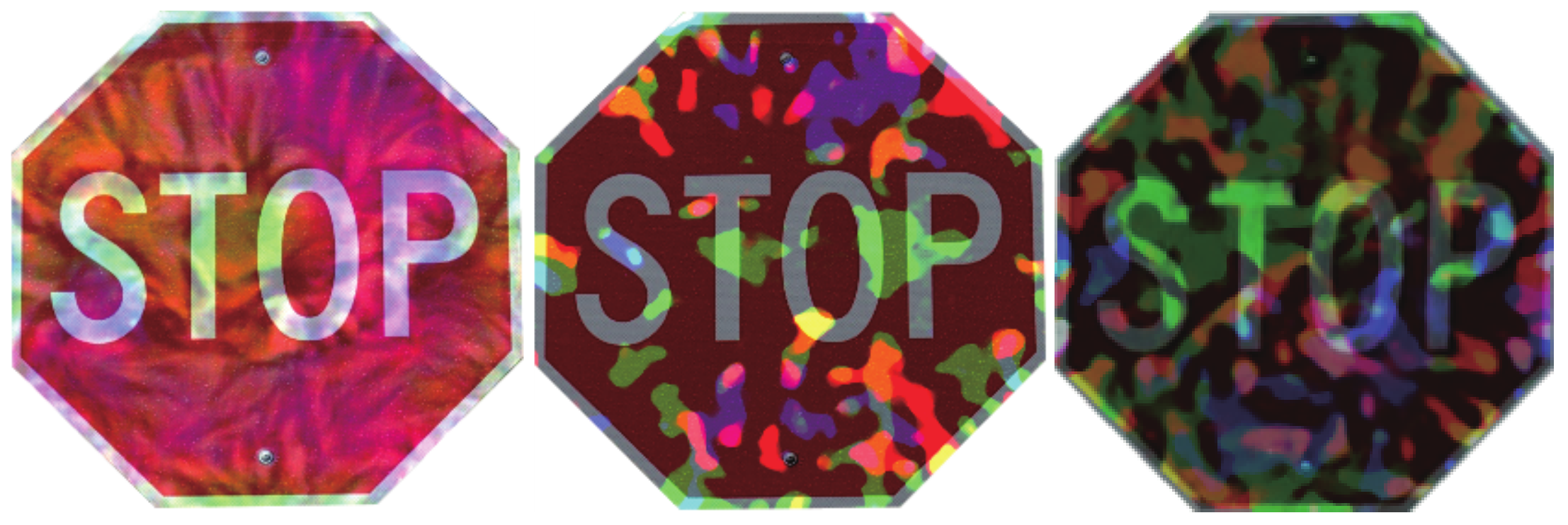}}
	\caption{We generate three adversarial stop signs with our attacking method. The first stop sign does not use $L_2$ distance penalty, 
and use termination criterion of successfully attacking 90\% of validation images. The second stop sign uses $L_2$ distance penalty in the objective function,
and terminates when 90\% of validation images are attacked. The last stop sign also adopts $L_2$ distance penalty, but performs
a  large fixed number of iterations. All three patterns reliably fool detectors when mapped into videos. However, physical instances of these patterns
are not equally successful. The first two stop signs, as physical objects, only occasionally fool Faster RCNN; the third one, which has a much more
extreme pattern, is more effective. }
	\label{fig:adv_stop}
\vspace{-2ex}
\end{figure}

\begin{figure*}
	\centerline{\includegraphics[width=1.0\linewidth]{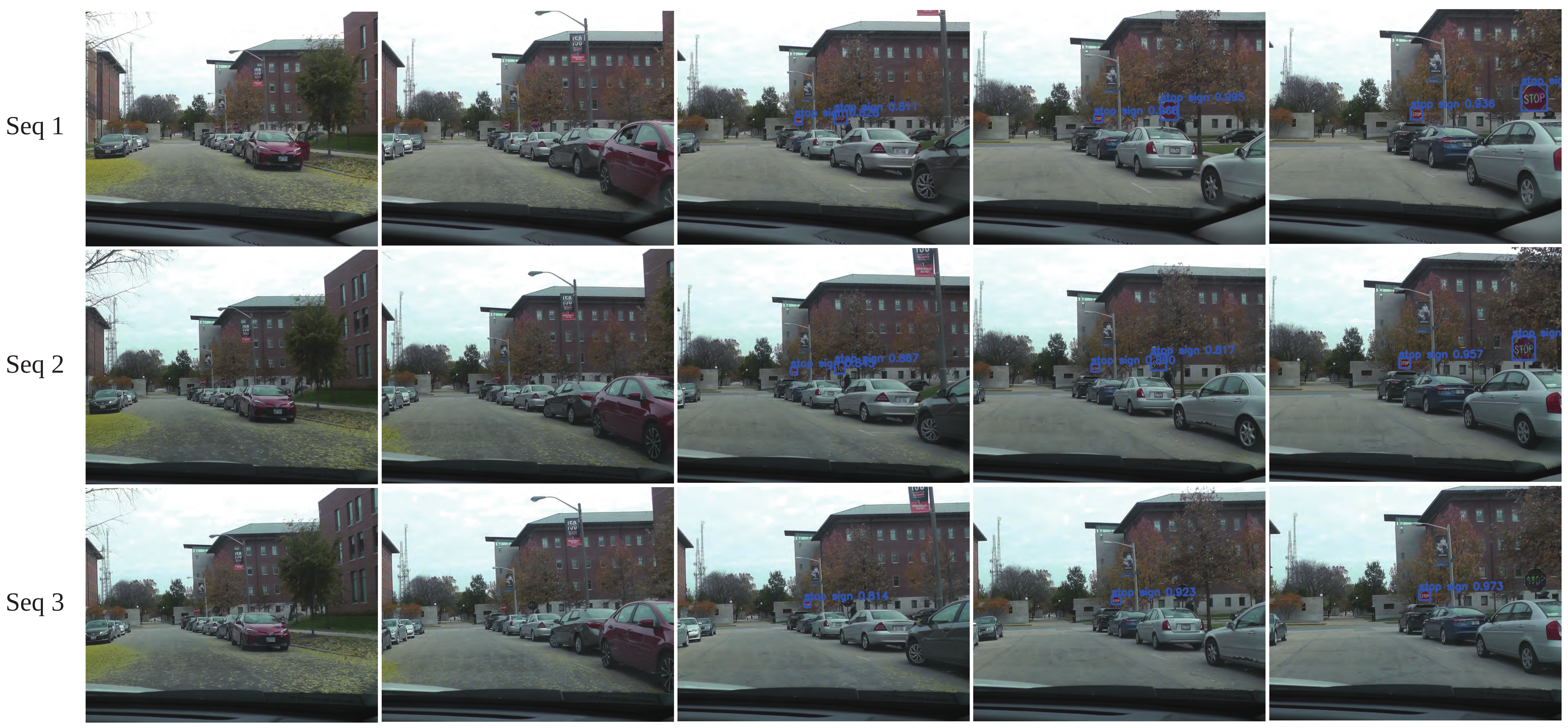}}
	\caption{We print the three adversarial stop signs from Figure~\ref{fig:adv_stop}, and stick them to a real stop sign. We took videos
while driving by these printed stop signs, and ran Faster RCNN on these videos. Notice that all of the adversarial perturbations
generalize well digitally. We only render the detection results for stop signs to make the figures clean.
The three sequences in this figure correspond to the three stop signs in order. In the first two sequences, stop signs
are detected without trouble, while in the last sequence, the stop sign is not detected in the video, so it is 
a physical adversarial stop sign. This may be the result of poor texture contrast against the tree, though this sequence was not seen in training.}
	\label{fig:physical_adv_stop}
\vspace{-2ex}
\end{figure*}

\subsection{Generalizing across view conditions}
\label{sec:multiple}

What we are really interested in is to produce a pattern that fails to be detected in any image. This is much harder because our pattern needs
to generalize to different view conditions and so on. We can still find adversarial patterns in this situation, but the patterns found by our process
involve significant changes of the stop signs and faces. 

{\bf Stop sign dataset}: we use a Panasonic HC-V700M HD camera to take 22 videos of the camera approaching stop signs, and extract 5 diverse frames from each video. 
Then we manually register all the stop signs, and use our attacking method to generate a unified adversarial perturbation for all the frames. 
We use 12 videos for generating adversarial perturbations (training), 5 videos for validation, and 5 videos for evaluation (testing). 
We use the validation set termination criteria described in Section~\ref{sec:attacking_stop_signs}.
Figure~\ref{fig:stop_multiple} gives an example video sequence, and its corresponding attacked video sequence. Table~\ref{table:digital_stop}
shows the stop sign detection rates in different circumstances. We plan to release the labelled dataset. 

{\bf Face dataset:}: we use a SONY a7 camera to take 5 videos of a still face from different distances and angles, and extract 20 diverse frames from each video. We
use the morphable face model approach to register all the faces, and use our attacking method to generate a unified adversarial perturbation. 
We use 3 videos for generating adversarial perturbations (training), 1 video for validation, and 1 video for evaluation (testing). 
Again, we use the validation set termination criteria described in Section~\ref{sec:attacking_stop_signs}.
Figure~\ref{fig:face_multiple} shows an example video sequence, and its corresponding attacked video sequence. In our experiments, this is the smallest
perturbations on faces that could generalize. Table~\ref{table:digital_face}
shows the face detection rates in different circumstances. Also, we plan to release the processed dataset. 

In summary, it is possible to attack stop signs and faces from multiple images, and require them to generalize to new similar view condition images. 
However, both of them require strong perturbation patterns to generalize. Refer to supplementary materials for details. 

\begin{table*}[h!]
	\begin{center}
		\resizebox{1.0\textwidth}{!}
		{
			\begin{tabular}{  c  c  | c  c  c  |  c   c   c | c  c  c   }
			 &  & test - far & test - medium & test - near & train - far & train - medium & train - near & val - far & val - medium & val - near \\
			\hline
			\hline
			 \multirow{2}{*}{Tree bg - L} & Stop1 & 0/4 ; 4/4 & 0/4 ; 1/4 & 0/2 ; 2/2 & 0/10 ; 6/10 & 0/10 ; 1/10 & 0/5 ; 5/5 & 1/4 ; 3/4 & 0/4 ; 3/4 & 0/2 ; 2/2 \\
			  &  Stop2 & 0/4 ; 4/4 & 1/4 ; 3/4 & 0/2 ; 2/2 & 2/10 ; 1/10 & 1/10 ; 0/10 & 2/5 ; 3/5 & 0/4 ; 1/4 & 0/4 ; 0/4 & 0/2 ; 0/2 \\
			\hline
			 \multirow{1}{*}{Tree bg - EL} & Stop3 & n/a ; n/a & n/a ; n/a & n/a ; n/a & 1/5 ; 5/5 & 0/5 ; 0/5 & 0/4 ; 1/4 & n/a ; n/a & n/a ; n/a & n/a ; n/a \\
			\hline
			\hline
			 \multirow{2}{*}{ Sky bg - L} & Stop1 & 0/6 ; 6/6 & 0/6 ; 5/6 & 0/3 ; 3/3 & 0/14 ; 13/14 & 0/14 ; 12/14 & 0/7 ; 6/7 & 0/6 ; 6/6 & 0/6 ; 5/6 & 0/3 ; 2/3 \\
			   & Stop2 & 0/6 ; 6/6 & 0/6 ; 6/6 & 0/3 ; 3/3 & 0/14 ; 13/14 & 2/14 ; 12/14 & 1/7 ; 6/7 & 0/6 ; 6/6 & 0/6 ; 5/6 & 1/3 ; 2/3 \\
			\hline
			 \multirow{1}{*}{Sky bg - EL} & Stop3 & 0/4 ; 3/4 & 0/4 ; 1/4 & 0/4 ; 3/4 & 0/5 ; 5/5 & 0/5 ; 5/5 & 0/4 ; 4/4 & n/a ; n/a & n/a ; n/a & n/a ; n/a \\
			\end{tabular}
		}
		\vspace{0.5ex}
		\caption{This table reports the detection rates of Faster RCNN and YOLO for the multiple image digital attacks of stop signs. 
In each cell,  the ratio before semicolon represents the detection rate for Faster RCNN, and the ratio after semicolon represents the detection 
rate for YOLO. Tree bg means the background of the stop sign is tree and has low contrast, and the Sky bg means the background of the stop sign 
is sky and has high contrast. L following the background means the perturbations are large, and EL means the perturbations are extremely large. 
We have three dark stop signs, and the detection rates are calculated at three different distances (far/medium/near)
on train/val/test splits. We can attack Faster RCNN with multiple view conditions, and the adversarial perturbations generalize to new
view conditions. The adversarial examples also generalize to YOLO, especially when the background is tree.}
		\label{table:digital_stop}
		\vspace{-2ex}
	\end{center}
\end{table*}

\begin{table*}[h!]
	\begin{center}
		\resizebox{1.0\textwidth}{!}
		{
			\begin{tabular}{  c  | c  c  c  c  |  c  c   c   c | c  c  c  c   }
			 & \multicolumn{4} {c|} {test} & \multicolumn{4} {c|} {train} & \multicolumn{4} {c} {val} \\
			 & ft-far & ft-near & sd-far & sd-near & ft-far & front-near & sd-far & sd-near & ft-far & ft-near & sd-far & sd-near \\
			\hline
			\hline
			S100 small perturbation & 2/3 & 2/4 & 6/6 & 6/7 & 0/19 & 0/21 & 0/9 & 0/11 & 1/4 & 6/8 & 3/4 & 3/4\\
			S100 medium perturbation & 3/3 & 2/4 & 6/6 & 3/7 & 5/19 & 1/21 & 1/9 & 0/11 & 2/4 & 4/8 & 4/4 & 3/4 \\
			S100 large perturbation & 0/3 & 0/4 & 1/6 & 0/7 & 0/19 & 0/21 & 0/9 & 0/11 & 0/4 & 0/8 & 0/4 & 0/4 \\
			S15 large perturbation & 0/1 & n/a & 0/2 & n/a & 0/3 & 0/2 & 0/2 & 0/2 & 0/1 & n/a & 0/2 & n/a \\
			\end{tabular}
		}
		\vspace{0.5ex}
		\caption{This table reports the detection rates of Faster RCNN based face detector~\cite{jiang2017face} for multiple image digital attacks of faces. S100 means there are 100 images in the experiments, and S15 means there are 15 images. Ft means frontal face, and sd means side face. 
When small perturbations are applied, attacks on all the training images succeed, but do not generalize to the validation and testing images. 
Only when large perturbations are applied, the attacks generalize to different view conditions. }
		\label{table:digital_face}
		\vspace{-2ex}
	\end{center}
\end{table*}

\begin{table*}[h!]
	\begin{center}
		\resizebox{1.0\textwidth}{!}
		{
			\begin{tabular}{  c  c  | c   c | c  c | c  c }
			 &  &  far - adv & far - clean & medium - adv & medium - clean & near - adv & near - clean  \\
			\hline
			\hline
			 \multirow{4}{*}{Tree bg - L} & Dark-Stop1 & 2/17 ; 4/17 & 16/17 ; n/a & 11/12 ; 12/12 & 12/12 ; n/a & 8/8 ; 8/8 & 8/8 ; n/a \\
			  &  Dark-Stop2 & 4/17 ; 5/17 & n/a ; n/a & 6/11 ; 10/11 & n/a ; n/a & 0/14 ; 12/14 & n/a ; n/a \\
			  &  Bright-Stop1 & 10/17 ; 16/17 & 17/17 ; n/a & 15/15 ; 15/15 & 15/15 ; n/a & 12/12 ; 12/12 & 12/12 ; n/a \\
			  &  Bright-Stop2 & 1/17 ; 2/17 & 14/17 ; n/a & 10/12 ; 12/12 & 12/12 ; n/a & 5/8 ; 8/8 & 8/8 ; n/a \\
			\hline
			 \multirow{2}{*}{Tree bg - EL} & Dark-Stop3 & 0/17 ; 0/17 & 17/17 ; n/a & 4/14 ; 10/14 & 14/14 ; n/a & 8/11 ; 7/11 & 11/11 ; n/a \\
			  & Bright-Stop3 & 0/17 ; 0/17 & 11/17 ; n/a & 1/11 ; 3/11 & 11/11 ; n/a & 0/7 ; 1/7 & 7/7 ; n/a \\
			\hline
			\hline
			 \multirow{4}{*}{ Sky bg - L} & Dark-Stop1 & 1/19 ; 5/19 & 0/19 ; n/a & 9/11 ; 11/11 & 0/13 ; n/a & 16/16 ; 16/16 & 14/16 ; n/a \\
			   & Dark-Stop2 & 0/25 ; 14/25 & 2/25 ; n/a & 5/15 ; 11/15 & 1/15 ; n/a & 24/24 ; 11/24 & 22/24 ; n/a \\
			   & Bright-Stop1 & 5/26 ; 23/26 & 0/26 ; n/a & 10/11 ; 11/11 & 1/11 ; n/a & 21/21 ; 21/21 & 14/21 ; n/a \\
			   & Bright-Stop2 & 1/23 ; 16/23 & 0/23 ; n/a & 10/11 ; 9/11 & 0/11 ; n/a & 20/24 ; 19/24 & 18/24 ; n/a \\
			\hline
			 \multirow{2}{*}{Sky bg - EL} & Dark-Stop3 & 14/27 ; 16/27 & 20/27 ; n/a & 3/13 ; 11/13 & 13/13 ; n/a & 26/27 ; 26/27 & 27/27 ; n/a \\
			   & Bright-Stop3 & 0/28 ; 11/28 & 0/24 ; n/a & 0/13 ; 10/13 & 2/13 ; n/a & 22/24 ; 10/24 & 18/24 ; n/a \\
			\end{tabular}
		}
		\vspace{0.5ex}
		\caption{The detection rates of the physical adversarial stop signs and physical clean stop signs with Faster RCNN and YOLO in different circumstances. The table layout is similar to Table~\ref{table:digital_stop}.
We have two stop signs with different brightness for large perturbations, and one stop sign with different brightness
for extremely large perturbations. We report both detection rates for 30 x 30 inches adversarial stop signs (adv) and 20 x 20 inches clean normal 
stop signs when applicable (clean).}
		\label{table:physical}
		\vspace{-2ex}
	\end{center}
\end{table*}

\subsection{Generalizing to the physical world}
There is a big gap between attacks in the digital world and attacks in the physical world, which means the adversarial perturbations
that generalize well in the digital world may not generalize to the physical world. We suspect this gap is 
due to various practical concerns, such as sensor properties, view conditions, printing errors, lighting, etc. 
In this paper, we print stop signs and perform physical experiments with them, but we believe similar conclusions apply to faces. 

We performed physical experiments with three adversarial perturbation patterns in Figure~\ref{fig:adv_stop}. Our results 
in Table~\ref{table:physical} show that the two less perturbed stop signs can still be detected by Faster RCNN, while the one with large perturbations
is hard to detect. The frames for physical experiments could be found in Figure~\ref{fig:physical_adv_stop}. Refer to our supplementary materials for videos. 

We performed analysis with the data from Table~\ref{table:digital_stop} and Table~\ref{table:physical}. $L_1$ regularized logistic regression 
is used to predict the success of our many different cases. The most important variable is detector (generalization from Faster RCNN to YOLO
is not strong); then whether the adversarial example is physical or not (digital attacks are more effective than physical); then scale (it is hard
to make a detector to miss a nearby stop sign). 

\subsection{Generalizing to YOLO}
\begin{figure}
	\centerline{\includegraphics[width=1.0\linewidth]{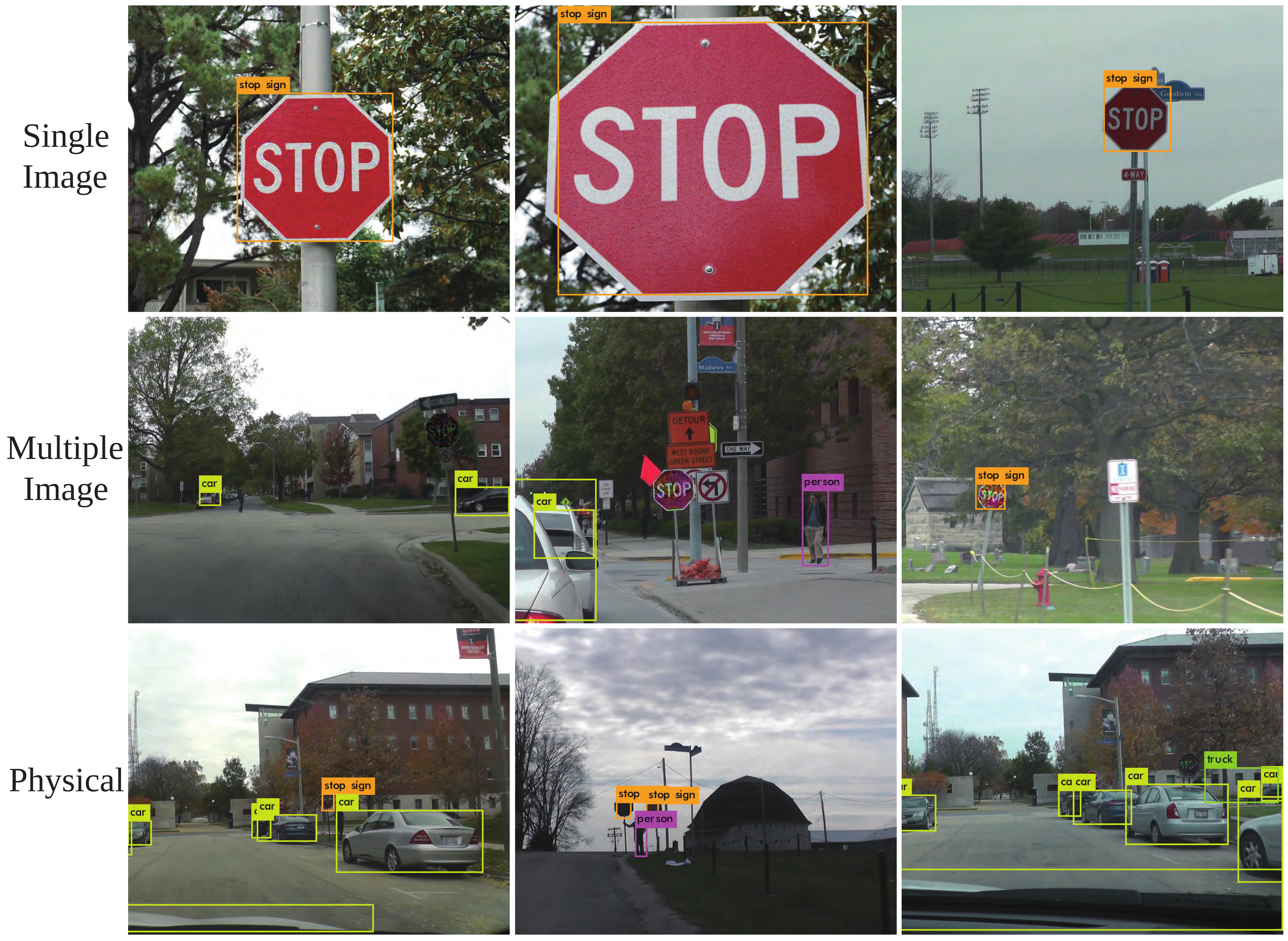}}
	\caption{We test whether adversarial examples generated from Faster RCNN generalize to YOLO. In the first row, these
adversarial examples are generated for a single image with small perturbations. YOLO can detect these stop signs without trouble. 
In the second row, these adversarial examples are generated from multiple images, and the digitally perturbed images can fool
YOLO in about half of the times. In the last row, physically printed adversarial stop signs can still fool YOLO in some circumstances.
The detailed summary can be found in Table~\ref{table:digital_stop} and Table~\ref{table:physical}. }
	\label{fig:yolo}
\end{figure}

Adversarial examples for a certain classifier generalize across different classifiers. To test out whether adversarial examples for 
Faster RCNN generalize across detectors, we feed these images into YOLO. 
We categorize our adversarial examples into three categories: single image examples with small perturbations, 
multiple image examples with large perturbations that generalize across viewing conditions digitally, 
and physical examples with large perturbations. Our experiments show that small perturbations do not
generalize to YOLO, while obvious perturbation patterns can generalize to YOLO with good probability. Examples are given in
Figure~\ref{fig:yolo}, and detection rates can be found in Table~\ref{table:physical} and Table~\ref{table:digital_stop}.

\begin{figure}
	\centerline{\includegraphics[width=1.0\linewidth]{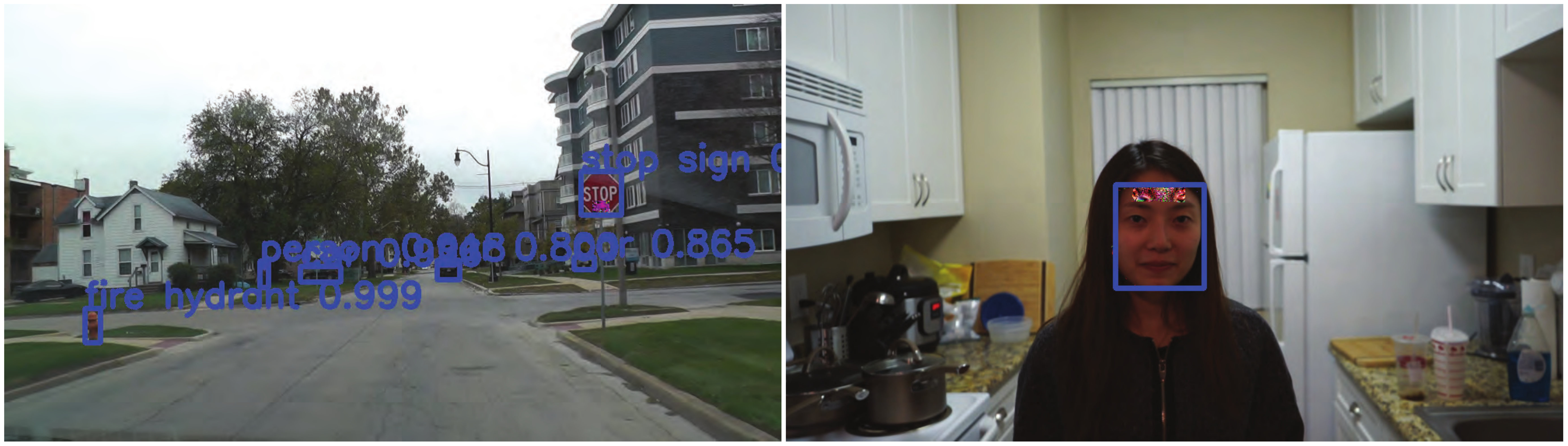}}
	\caption{Localized attacks for stop signs and faces fail in the multiple view condition setting. We applied attacks on regions of the
stop signs and faces with very large number of iterations, and introduced extremely large perturbations, but the objects are still detected. 
In detail, localized attacks on stop signs can sometimes digitally fool far stop signs, but not for middle and near stop signs; localized attacks
on faces cannot fool face detector. The first image is an example of perturbed stop signs, and the second image is an example of
perturbed faces. }
	\label{fig:attack_region}
\end{figure}

\subsection{Localized attacks fail}
In previous settings, we attack the whole masked objects in the images, however, it is usually hard to apply such attacks in the physical world. 
For example, modifying the whole stop sign patterns is useless in practice, and wearing a whole face mask with perturbation patterns is hard too. 
It would be more effective attack if one can manufacture small stickers with perturbation patterns, and when the sticker is attached to a small region 
of the stop sign or the face forehead, the detector would fail. 
Evtimov {\em et al.}~\cite{evtimov2017robust} showed an example that successfully attacked stop sign classifiers. 
We try to generate adversarial patterns constrained to a fixed region of the objects to fool detectors, however, we find these attacks
only occasionally successful (stop signs) or wholly unsuccessful (faces). Figure~\ref{fig:attack_region} shows some examples. 

\subsection{Simple defenses fail}
\begin{figure}
	\centerline{\includegraphics[width=1.0\linewidth]{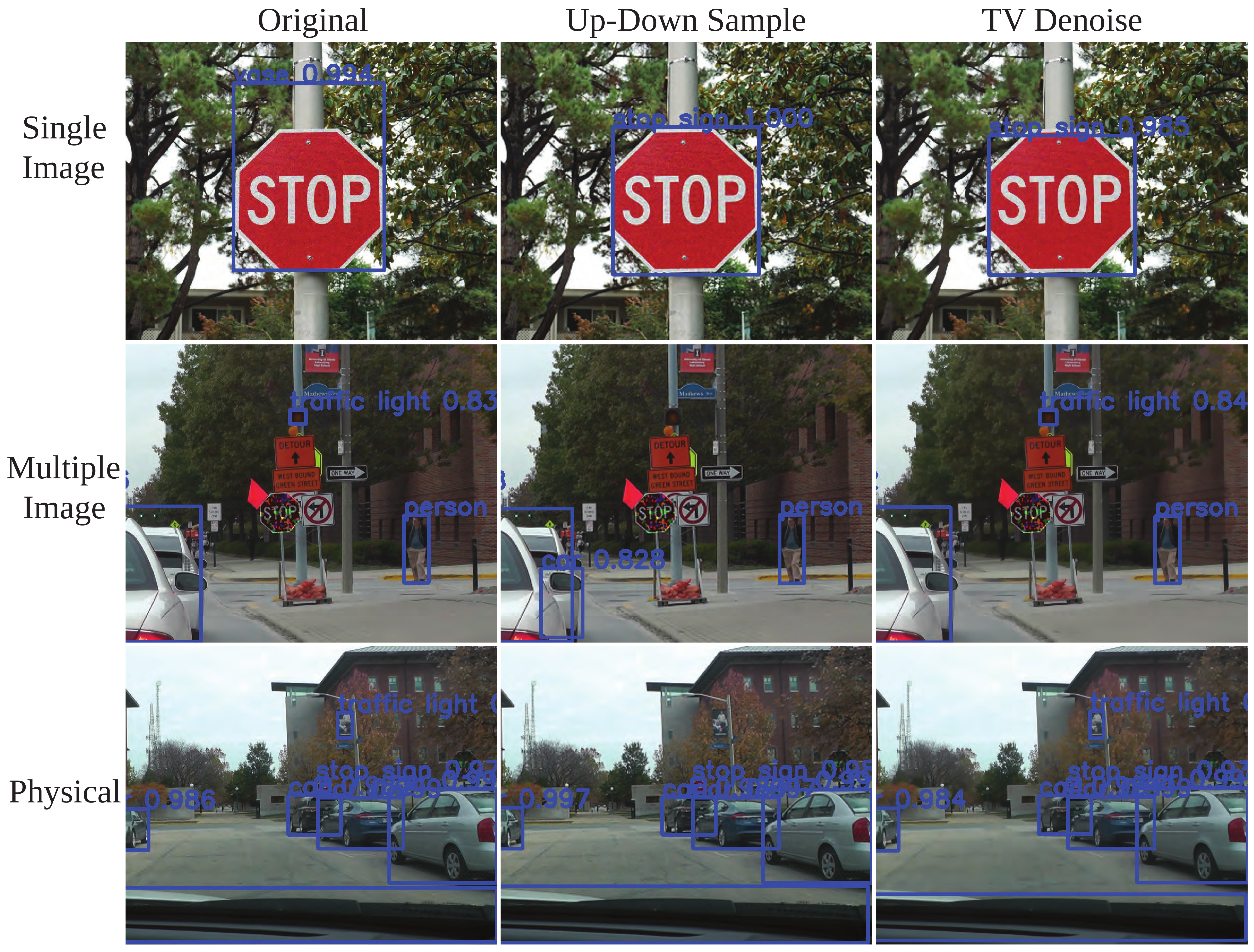}}
	\caption{We apply simple defenses~\cite{guo2017countering} to adversarial examples generated for Faster RCNN. Up-Down Sample 
means down sample the image resolution by half, and then upsample it to the original resolution. TV Denoise means denoising the image with
total variation regularization, which will remove the high frequency information and keep the low frequency information. In the first row,
the adversarial examples generated from a single image with small perturbations can be detected after simple image processing. In the second
row, the adversarial examples generated from multiple view conditions still cannot be detected after simple defense. In the last row, the
physically printed adversarial stop signs still cannot be detected after simple defense. }
	\label{fig:defense}
\end{figure}

Recently, Guo {\em et al.}~\cite{guo2017countering} showed that simple image processing could defeat a majority of imperceivable adversarial attacks. 
We assume detectors should run at frame rate, so exclude image quilting. We investigated down-up sampling and total variation smoothing defense.
We find that these methods can defeat attacks on a single image, but cannot disrupt the large patterns needed to produce adversarial examples
that generalize,  see Figure~\ref{fig:defense} and Table~\ref{table:defense}.
Our hypothesis for this phenomenon is that tiny perturbations work with numerical accumulation mechanism, which is not robust to changes, while 
obvious perturbations work with pattern recognition mechanism, which is more robust and can better generalize. 

\begin{table}
	\begin{center}
		\resizebox{0.47\textwidth}{!}
		{
			\begin{tabular}{  c  c | c  c  c  c }
			 & & NonAttack & Adversarial & UP & TV \\
			\hline
			\hline
			Single & & 10/10 & 0/10 & 10/10 & 10/10 \\
			\hline
			\multirow{3}{*}{Multiple} & stop1 & 110/110 & 1/110 & 8/110 & 8/110 \\
			& stop2 & 110/110 & 10/110 & 18/110 & 9/110 \\
			& stop3 & 40/40 & 1/40 & 1/40 & 2/40 \\
			\hline
			\multirow{3}{*}{Physical} & stop1 & 109/187 & 121/185 & 118/185 & 116/185 \\
			& stop2 & 77/159 & 86/201 & 84/201 & 88/201 \\
			& stop3 & 151/205 & 78/209 & 72/209 & 85/209 \\
			\end{tabular}
		}
		\vspace{0.5ex}
		\caption{We evaluate the effectiveness of simple defense methods. UP means downsample the input image resolution by half, and then
upsample it to the original size. TV means total variation denoise, which removes high frequency information. The physical non attack numbers are
counted from a real stop sign near our adversarial one. These simple defense methods are effective for single image perturbations, but not effective for 
multiple image perturbations that can generalize. They also cannot defeat physical adversarial perturbations. }
		\label{table:defense}
		\vspace{-2ex}
	\end{center}
\end{table}

\section{Conclusion}
We have demonstrated the first adversarial examples that can fool detectors. Our construction yields physical objects that
fool detectors too. However, all the adversarial perturbations we have been able to construct require large perturbations. This suggests that
the box prediction step in a detector acts as a form of natural defense. We speculate that better viewing models in our construction may yield
a smaller gap between physical and digital results. Our patterns may reveal something about what is important to a detector.

{\small
\bibliographystyle{ieee}
\bibliography{egbib}
}

\end{document}